\documentclass[letterpaper,10pt,conference]{ieeeconf}

\IEEEoverridecommandlockouts
\usepackage[utf8]{inputenc}
\usepackage{amsmath}
\usepackage{amssymb}
\usepackage{array}
\usepackage{tabularx}
\usepackage{booktabs}
\usepackage{graphicx}
\usepackage{wrapfig}
\usepackage{bm}
\usepackage{needspace}
\makeatletter
\def\endfigure{\end@float}
\def\endtable{\end@float}

\let\@IEEEoriginalmakecaption\@makecaption
\long\def\@makecaption#1#2{%
    \ifx\@captype\@IEEEtablestring
        {\centering\footnotesize #1:~#2\par}%
        \@IEEEtablecaptionsepspace%
    \else
        \@IEEEoriginalmakecaption{#1}{#2}%
    \fi
}
\def\@IEEEtablecaptionsepspace{\vskip 0.25\baselineskip\relax}%
\makeatother
\usepackage[caption=false,font=footnotesize]{subfig}
\usepackage{capt-of}

\newcommand{\meanstd}[2]{\ensuremath{#1\,{\scriptstyle\mathbin{\pm}#2}}}
\newcommand{\boldmeanstd}[2]{\ensuremath{\bm{#1}\,{\scriptstyle\mathbin{\pm}#2}}}

\title{\LARGE\bfseries ParkourFormer: Integrating Predictive Supervision and Sequence Modeling into Parkour Locomotion}

\author{
  \authorblockN{Yanheng Mai$^{1,2,*}$, Wenhao Xu$^{1,2,*}$, Zirui Huang$^{2,*}$, Kailun Huang$^{1,*}$, Yifei Fu$^{2}$, \\
  Xinjue Wang$^{2}$, Shengwei Dong$^{2}$ , Yanzhe Xie$^{1}$, Renjing Xu$^{1,\dagger}$}\\
  \authorblockA{$^{1}$The Hong Kong University of Science and Technology (Guangzhou), $^{2}$CLAI-LAB, CL-TECH \\
  $^{*}$ Equal Contribution $\dagger$ Corresponding author}
}

\begin{document}
\maketitle
\thispagestyle{empty}
\pagestyle{empty}

\begin{abstract}
Humanoid parkour requires locomotion policies to coordinate whole-body dynamics across rapidly changing terrains such as stairs, gaps, slopes, and obstacles. Existing learning-based humanoid locomotion policies typically infer actions from current and historical observations, while the future evolution of the robot state is represented only implicitly. Explicitly anticipating short-horizon state evolution is particularly beneficial for multiple terrains locomotion, where current actions must account for upcoming contact transitions and whole-body dynamics. We present \textbf{ParkourFormer}, a Transformer-based sequence modeling framework that reformulates humanoid locomotion as a future-conditioned decision-making problem. The current robot state queries historical sensorimotor trajectories through cross-attention, while a lightweight prediction head forecasts short-horizon future proprioceptive states. The future states are predicted solely from historical observations and actions, with ground-truth future states used only as supervision targets, and are fused with temporal features to generate the current action without future-information leakage. We evaluate ParkourFormer on a diverse multi-terrain humanoid parkour benchmark including stairs, gaps, slopes, rough terrain, and obstacle traversal. ParkourFormer achieves a \textbf{93.96\%} mean traversal success rate, compared with \textbf{90.49\%} for a vanilla Transformer and \textbf{87.16\%} for a MoE-MLP policy, while maintaining a single policy across all terrain types. These results demonstrate that explicit future-state modeling significantly improves robustness and generalization for agile whole-body locomotion.
\end{abstract}




\section{Introduction}

\begin{figure*}[t]
    \centering
    \includegraphics[width=\textwidth]{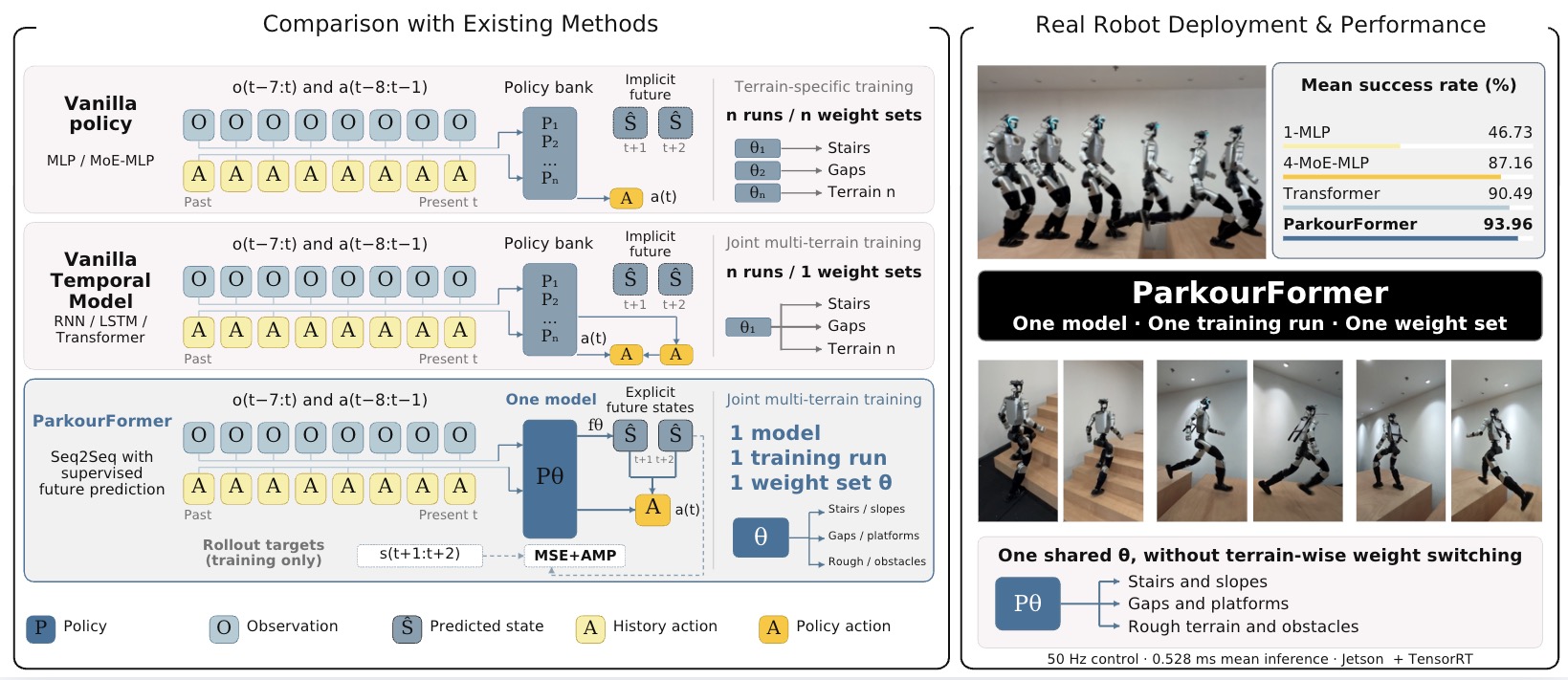}
    \caption{
    \textbf{Comparison with existing paradigms for multi-terrain humanoid locomotion.}
    Earlier MLP/MoE-MLP-based methods often rely on terrain-specific training, multiple weight sets, and FSM-style policy switching due to limited temporal representation capacity.
    Vanilla temporal models improve sequence modeling and enable joint multi-terrain training with a single shared policy.
    ParkourFormer further introduces explicit short-horizon future prediction and combines it with predictive supervision and AMP, yielding a single model with one training run and one weight set, while achieving stronger success rates and better generalization across diverse terrains.
    }
    \label{fig:parkourformer_overview}
\end{figure*}

Legged robots, especially humanoids, have made substantial progress from stable locomotion on flat ground toward agile traversal of complex environments through reinforcement learning (RL), sim-to-real transfer, and whole-body control~\cite{HeT-RSS-25,hwangbo2019learning,gu2024humanoid}. 
Recent learning-based policies have demonstrated strong locomotion capabilities over challenging terrains~\cite{choi2023learning,van2024revisiting,zhang2024learning}. 
Parkour, however, places stronger demands on a unified policy: the robot must continuously switch among different motion patterns when traversing stairs, gaps, slopes, platforms, and irregular obstacles while maintaining whole-body balance. 
Such heterogeneous terrain transitions require the policy not only to represent diverse locomotion behaviors, but also to exploit their temporal structure for online control.

Legged locomotion is inherently sequential, as contact states, momentum, and joint trajectories evolve continuously and couple current control decisions with preceding robot states and subsequent motion evolution~\cite{2006The,2023BiConMP,2017Robust,kim2020dynamic}. 
This motivates temporal models that can capture long-range dependencies and different motion patterns within a unified representation~\cite{sleiman2021unified,winkler2018gait}. 
Recurrent architectures such as LSTM~\cite{hochreiter1997long} and RNN~\cite{elman1990finding} provide temporal memory, while Transformers~\cite{vaswani2017attention} offer attention-based sequence modeling that can selectively retrieve relevant information from historical trajectories. 
Decision Transformer further shows that decision making can be formulated through causal sequence modeling~\cite{chen2021decision}. 
For multi-terrain locomotion, such a formulation provides a flexible way to model heterogeneous temporal motion representations within a single policy rather than relying only on instantaneous observations.

However, existing temporal locomotion policies typically use historical observations to directly infer actions, leaving future robot-state evolution represented only implicitly in the learned features. 
We therefore formulate humanoid locomotion as a \emph{causal sequence-to-sequence prediction and control problem}: historical observations and actions form the causal context, the current robot state queries this temporal history, and a lightweight prediction head explicitly predicts short-horizon future proprioceptive states. 
The prediction is strictly causal: ground-truth future states are used only as supervision targets during training and are never provided as policy inputs, preventing shortcut learning or future-information leakage. 
The predicted states are subsequently fused with the temporal representation and directly condition current action generation, allowing the policy to connect past motion patterns, current state estimation, and anticipated short-term dynamics within one sequence model. 
This predictive supervision also provides an explicit learning objective for shaping temporal representations, complementing reinforcement learning with future-state prediction~\cite{schwarzer2020data}.

\begin{figure}[b]
    \centering
    \includegraphics[width=\linewidth]{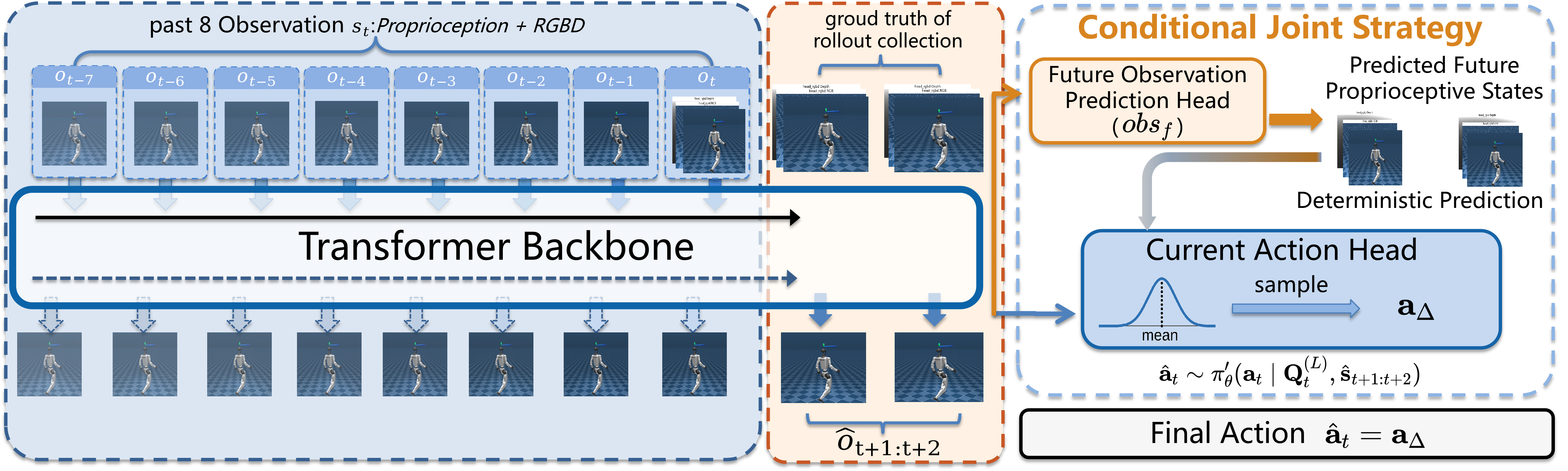}
    \caption{
        \textbf{Conditional Joint Strategy and Future Prediction Head.} 
        The future observation prediction head (\( \text{obs}_{f} \)) generates deterministic predictions of the next two proprioceptive states from the Transformer backbone features. 
        These predicted future observations are concatenated with the historical features and directly fed into the current action head to produce action deltas, enabling the current action to be jointly conditioned on both past observations and predicted future states.
    }
    \label{fig:parkourformer_joint_strategy}
\end{figure}

Based on this formulation, we propose \textbf{ParkourFormer}, a Transformer-based framework for unified multi-terrain humanoid parkour. 
Its attention backbone learns temporal representations shared across heterogeneous locomotion behaviors, while the ``past $\rightarrow$ current $\rightarrow$ future'' predictive structure explicitly models short-horizon state evolution for future-conditioned control. 
Unlike terrain-specific policies or architectures that introduce separate experts for different locomotion regimes, ParkourFormer maintains a single policy across diverse terrains. 
The overview of ParkourFormer is shown in Fig.~\ref{fig:parkourformer_overview}, with its future-conditioned policy illustrated in Fig.~\ref{fig:parkourformer_joint_strategy}. The contributions are summarized as follows:

\textbf{Causal Sequence-to-Sequence Modeling for Locomotion}: 
We formulate humanoid locomotion as a causal seq2seq problem in which the current state queries historical sensorimotor trajectories and explicitly predicts short-horizon future states, connecting past, current, and predicted future dynamics within a unified temporal model.

\textbf{Predictive Supervision and Future-conditioned Control}: 
We introduce supervised future-state prediction into policy learning and directly condition the action head on the predicted states. And this design provides explicit predictive supervision for learning temporal motion representations.

\textbf{Unified Multi-Terrain Motion Representation}: 
ParkourFormer jointly learns diverse locomotion behaviors across stairs, slopes, obstacles, and gaps in a single training process, and deploys one shared set of policy weights across all terrains, without finite-state-machine switching, terrain-specific policy selection, or separately trained controllers.


\section{Related Work}
\label{sec:related_work}

\begin{figure*}[t]
    \centering
    \includegraphics[width=\linewidth]{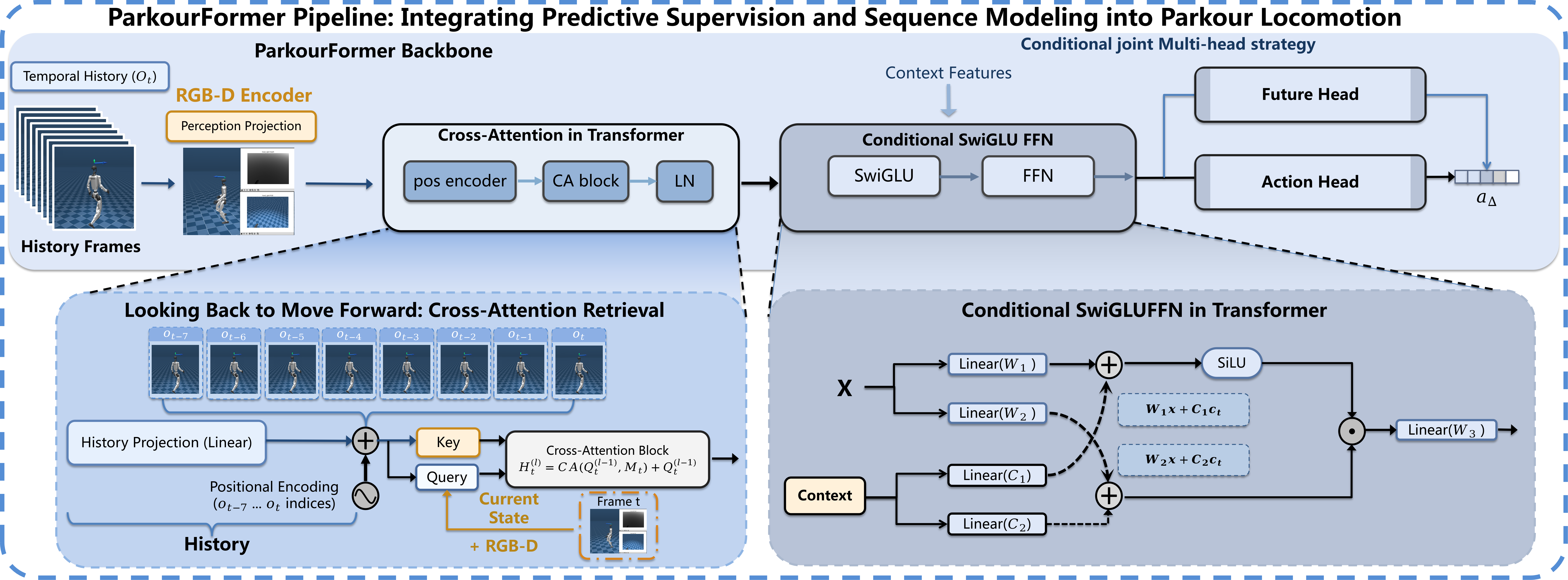}
    \caption{
        \textbf{ParkourFormer Seq2Seq locomotion pipeline.} 
        ParkourFormer is a Transformer-based Seq2Seq framework for parkour locomotion. It processes historical observations via cross-attention, fuses terrain features using conditional SwiGLU FFN, and employs an asymmetric critic for value estimation. The actor generates actions as a sequence-to-sequence task, supported by future prediction and multi-discriminator regularization. Training is performed with a novel joint PPO~\cite{schulman2017proximal} formulation that separately optimizes the future prediction and current action components.
    }
    \label{fig:parkourformer_pipeline}
\end{figure*}

\subsection{Learning-based Humanoid Locomotion and Parkour}

Deep reinforcement learning has enabled increasingly agile legged locomotion through actuator modeling, domain randomization, and sim-to-real transfer~\cite{hwangbo2019learning,gu2024humanoid,HeT-RSS-25}. 
Recent works further extend these capabilities to expressive whole-body behaviors~\cite{cheng2024expressive} and challenging terrain traversal. 
Robot Parkour~\cite{zhuang2023robot} introduces curriculum learning for agile obstacle traversal, while ANYmal Parkour~\cite{2024ANYmal} employs a hierarchical formulation to compose multiple locomotion skills. 
Extreme Parkour~\cite{cheng2024extreme} distills terrain-specific experts into a perceptive student policy, and recent humanoid systems such as Hiking in the Wild~\cite{zhu2026hiking} and PHP~\cite{wu2026perceptive} demonstrate perceptive parkour through MoE-based control and motion matching, respectively. 
Concurrent works further explore vision-aware behavior distillation and terrain-specific learning objectives~\cite{sun2026now}, as well as foothold-aware open-world humanoid traversal~\cite{yu2026ssr}. 
In contrast to terrain-level skill switching or separately trained terrain-specific controllers, our goal is to jointly learn heterogeneous terrain behaviors in one training process and deploy one policy weights.

\subsection{Multi-Terrain and Temporal Policy Modeling}

A key challenge in unified parkour is representing locomotion behaviors with substantially different contact dynamics within a single policy. Ziegltrum et al.~\cite{ziegltrum2026quadruped} apply sparsely gated MoE to perceptive parkour, while CMoE~\cite{ma2026cmoe} further encourages terrain-dependent expert specialization through contrastive learning. 
An alternative direction is to model heterogeneous behaviors within a shared temporal representation. 
Radosavovic et al.~\cite{radosavovic2024humanoid} formulate humanoid locomotion as causal next-token prediction using a Transformer, demonstrating the potential of sequence modeling for real-world control. 
Recent CReF~\cite{hao2026cref} similarly emphasizes richer temporal and cross-modal representations through proprioception-queried attention and recurrent integration.
ParkourFormer follows the sequence-modeling direction and further couples a shared Transformer representation with explicit future-state prediction. 
We formulate locomotion as a causal Seq2Seq process in which historical observations and actions provide the context for predicting short-horizon future proprioceptive states, which subsequently condition current action generation. 
This predictive objective encourages the Transformer to organize heterogeneous motion patterns within a shared temporal representation while retaining a single jointly trained policy across all terrains.

\section{Method}
\label{sec:Method}

\subsection{Transformer-based and temporal Policy Architecture}

At each control step $t$, ParkourFormer takes as input the policy observation $o_t \in \mathbb{R}^{96}$, the RGB-D frame $\mathbf{d}_t$, and the AMP-style motion state $s_t \in \mathbb{R}^{67}$~\cite{peng2021amp}. The overall policy architecture is illustrated in Fig.~\ref{fig:parkourformer_pipeline}. The observation $o_t$ and state $s_t$ are defined as:
\begin{equation}
    \begin{aligned}
    o_t &= \bigl[\, \boldsymbol{\omega}_a,\; \mathbf{g}_p,\; \mathbf{v}_c,\; \mathbf{q}_p,\; \mathbf{q}_v,\; \mathbf{a}_{t-1} \,\bigr], \\
    s_t &= \bigl[\, \mathbf{g}_p,\; \mathbf{q}_p,\; \mathbf{q}_v,\; \mathbf{v}_l,\; \boldsymbol{\omega}_a \,\bigr],
    \end{aligned}
    \label{eq:obs_and_state}
\end{equation}
where \(\boldsymbol{\omega}_a\in\mathbb{R}^3\) is the base angular velocity, \(\mathbf{g}_p\in\mathbb{R}^3\) is the projected gravity, \(\mathbf{v}_c\in\mathbb{R}^3\) is the velocity command, \(\mathbf{q}_p\in\mathbb{R}^{29}\) and \(\mathbf{q}_v\in\mathbb{R}^{29}\) are the joint positions and velocities, \(\mathbf{a}_{t-1}\in\mathbb{R}^{29}\) is the previous action, and \(\mathbf{v}_l\in\mathbb{R}^3\) is the base linear velocity.

\subsubsection{Temporal Modeling and RGB-D Input by Position Encoder}

The policy history is \(\mathbf{o}_t=\{o_{t-7},\ldots,o_t\}\). The current RGB-D frame is encoded as a depth token \(\mathbf{z}_t=\mathcal{F}_{\mathrm{RGB-D}}(\mathbf{d}_t)\in\mathbb{R}^{128}\). Historical observations are projected into token space and augmented with positional encodings, turning the frame stack into an ordered temporal context.

\subsubsection{Looking Back to Move Forward: Cross-Attention for WBC}

The temporal backbone uses cross-attention: the current observation and depth token form the query, while \(\mathbf{o}_t\) provides the key-value memory:
\begin{equation}
\mathbf{X}_t=\mathbf{W}_1\mathbf{o}_t\in\mathbb{R}^{8\times128},\qquad
\tilde{x}_t=\mathbf{W}_2[x_t;\mathbf{z}_t]\in\mathbb{R}^{128},
\end{equation}
where \(\mathbf{W}_1\) is the observation token projection, \(\mathbf{W}_2\) is the visual-observation fusion layer, \(x_t\) is the current observation token, and \(\mathbf{M}_t\in\mathbb{R}^{8\times128}\) is the positional memory. The initial query is \(\mathbf{Q}_t^{(0)}=\mathrm{LN}(W_q\tilde{x}_t)\in\mathbb{R}^{2\times128}\). Each encoder layer applies cross-attention and FFN residual updates:
\begin{equation}
\mathbf{H}_t^{(l)} =
\mathrm{CA}\!\left(\mathbf{Q}_t^{(l-1)}, \mathbf{M}_t\right)
+ \mathbf{Q}_t^{(l-1)},
\end{equation}
\begin{equation}
\mathbf{Q}_t^{(l)} =
\mathrm{FFN}\!\left(\mathrm{LN}(\mathbf{H}_t^{(l)})\right)
+ \mathrm{LN}(\mathbf{H}_t^{(l)}).
\end{equation}

\subsubsection{Conditional FFN in Transformer Actor-Critic}

After cross-attention, the query features $\mathbf{Q}_t^{(l)}$ are processed by a terrain-conditioned SwiGLU block. This module integrates the terrain context embedding $\mathbf{c}_t$, which is derived from the RGB-D encoder, into the feature transformation via multiplicative gating, thereby enabling geometry-adaptive modulation of intermediate representations. Formally, the ConditionalSwiGLU block computes:

\begin{equation}
\mathrm{FFN}({x}, c_t)
=
\mathbf{W}_5\!\left(
\mathrm{SiLU}\!\left(
\mathbf{W}_3{x} + C_1 c_t
\right)
\odot
\left(
\mathbf{W}_4{x} + C_2 c_t
\right)
\right),
\end{equation}

where $\mathbf{W}_3, \mathbf{W}_4, \mathbf{W}_5$ are learnable linear projections and $C_1, C_2$ denote context-conditioned modulation layers. This mechanism enables terrain-aware features to directly modulate intermediate Transformer activations via conditional gating.

\subsection{Supervised signal with future prediciton and AMP}
The predicted states \(\hat{\mathbf{s}}_{t+1:t+2}\) guide the action head and extend the AMP discriminator input. Let $\mathbf{s}_{t} = \{s_{t-7}, \ldots, s_t\}$ denote the real eight-frame AMP history. We further define $\tilde{\mathbf{s}}_{t} \in \mathbb{R}^{10 \times 67}$ as the complete discriminator sequence, and $\bar{\mathbf{s}}_{t} \in \mathbb{R}^{10 \times 67}$ as the reference action-state sequence.
                                                              
\subsubsection{Additional Supervised Signal for the future prediction head}
The prediction head is optimized jointly with PPO using two-step rollout supervision, while invalid episodes in the rollout are masked out.
\begin{equation}
    \mathcal{L}_{\mathrm{pred}}
    =
    \frac{1}{|\mathcal{M}_{\mathrm{pred}}|}
    \sum_{i}\sum_{k=1}^{2}
    m_{i,k}\,
    \left\|\hat{s}_{i,k}-s_{i,k}\right\|_2^2.
\label{eq:pred_loss}
\end{equation}
Here, \(s_{i,k}\) is the rollout target, \(\hat{s}_{i,k}\) is the prediction, and \(m_{i,k}\in\{0,1\}\) is the validity mask. \(\mathcal{L}_{\mathrm{pred}}\) constraint enforces the prediction head to match the kinematic distribution.
                                                                      
\subsubsection{Future Proprioception Prediction in Discriminator Input}
To ensure that the predicted state remains on the motion manifold defined by the reference AMP data, it is appended to the real AMP history data. This allows the discriminator to access a continuous sequence containing both observed and expected motions, rather than relying solely on the supervised loss:
\begin{equation}
\tilde{\mathbf{s}}_{t}
=
\left[\mathbf{s}_{t};\hat{\mathbf{s}}_{t+1:t+2}\right].
\end{equation}

\begin{figure}[t]
    \centering
    \includegraphics[width=\linewidth]{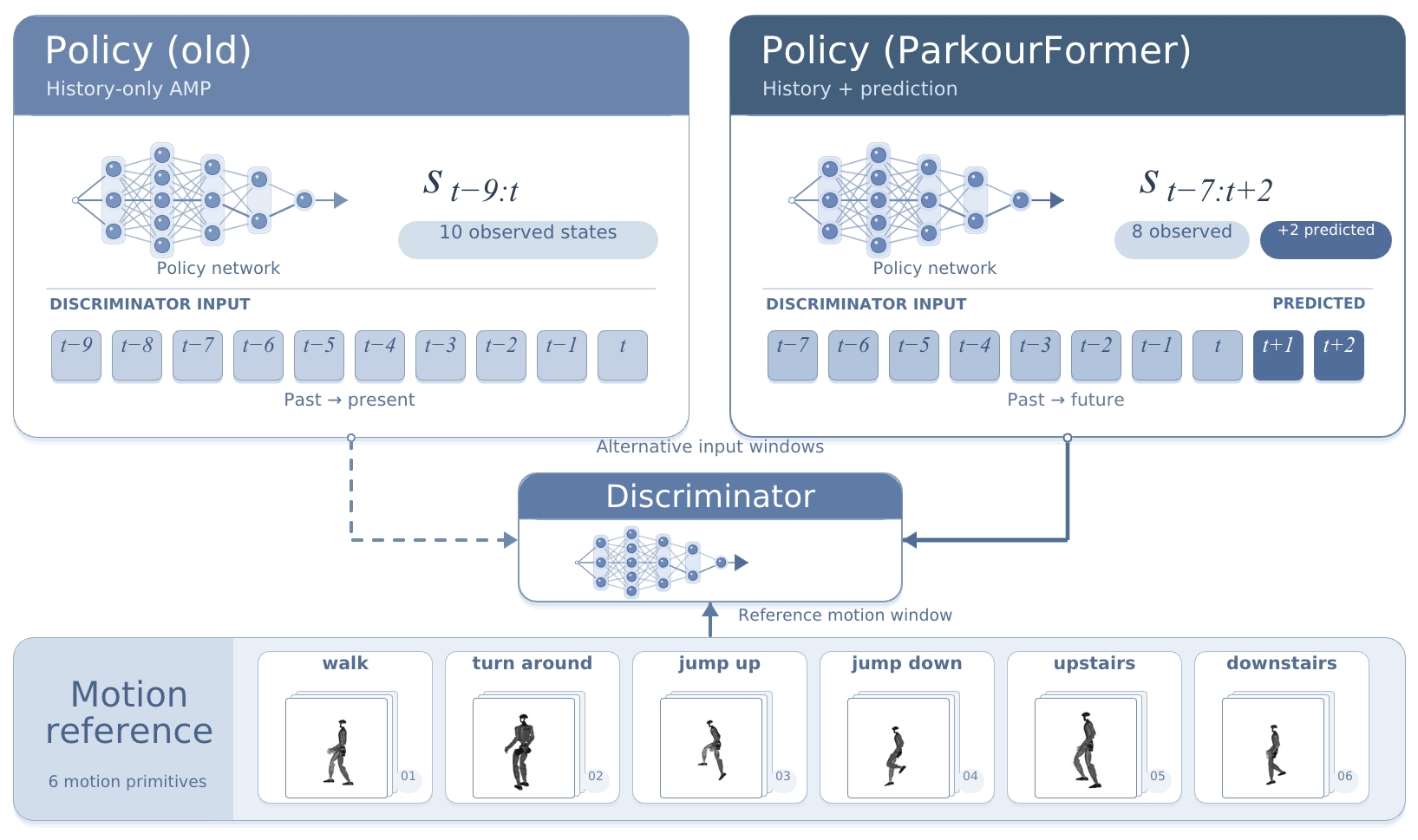}
    \caption{\textbf{Comparison of discriminator inputs.} ParkourFormer augments the AMP history with predicted future states.}
    \label{fig:compare}
\end{figure}

\subsection{Model Training and Loss Function}
\label{sec:3.3}

1) \textbf{Action Space}: The policy action $\mathbf{a}_t\in\mathbb{R}^{29}$, an action delta around the nominal pose, is converted to PD-tracked joint targets and executed by the low-level controller.

2) \textbf{Policy Network}: The current fused feature queries the historical tokens, and the prediction head forecasts two future AMP-style states:
\begin{equation}
\hat{\mathbf{s}}_{t+1:t+2} = f_\theta({\mathbf{s}}_{t}, \mathbf{Q}_t^{(L)}),
\end{equation}
\begin{equation}
\hat{\mathbf{a}}_{t} \sim {\pi_\theta'}({\mathbf{a}}_{t} \mid \mathbf{Q}_t^{(L)}, \hat{\mathbf{s}}_{t+1:t+2}),
\end{equation}

We model future state prediction via the branch $f_\theta$, and condition the action branch $\pi_\theta'$ on the predicted future states for decision making. Both branches share a common backbone with separate output heads.



3) \textbf{Value Network}: The asymmetric critic uses privileged simulation observations to improve value estimation, where privileged observations comprise all actor observations augmented with \(\mathbf{v}_l\).

4) \textbf{Reward Function and Training}: The total reward combines task-specific reward terms with an adversarial motion prior:
\begin{equation}
    R_{\mathrm{total}} = R_{\mathrm{task}} + R_{\mathrm{AMP}}.
\end{equation}
The task reward is implemented as a weighted sum over the complete set of task-reward terms:
\begin{equation}
    R_{\mathrm{task}} = \Delta t\sum_{k\in\mathcal{K}_{\mathrm{task}}} w_k r_k,
    \label{eq:task_reward}
\end{equation}
where $\mathcal{K}_{\mathrm{task}}$ is the full task-reward index set, $r_k$ is the unweighted output of term $k$, and $w_k$ is its configured coefficient. Table~\ref{tab:reward_functions} reports a core subset $\mathcal{K}_{\mathrm{core}}\subsetneq\mathcal{K}_{\mathrm{task}}$ and is not an exhaustive listing of the implemented task reward. The factor $\Delta t$ follows the time-step scaling used in the implementation. In the table, $\mathbb{I}_{\mathrm{move}}=[\lVert\mathbf{v}_{xy}^c\rVert_2>0.15\lor\lvert\omega_z^c\rvert>0.15]$ and $\delta_j=\lVert\Delta\mathbf{p}_j\rVert_2$.

\begin{table}[tb]
\centering
\caption{\textbf{Core task-reward terms.} Selected reward components and configured coefficients used in training.}
\label{tab:reward_functions}
\setlength{\tabcolsep}{1.5pt}
\scriptsize
\renewcommand{\arraystretch}{1.08}
\begin{tabularx}{\linewidth}{@{}>{\raggedright\arraybackslash}p{0.25\linewidth}>{\raggedright\arraybackslash}X>{\centering\arraybackslash}p{0.11\linewidth}@{}}
\toprule
\textbf{Term} & \textbf{Equation} & \textbf{Weight} \\
\midrule
planar velocity tracking & $\exp\!\left(-\lVert\mathbf{v}_{xy}^{c}-\mathbf{v}_{xy}^{b}\rVert_2^2/\sigma^2\right),\ \sigma=0.5$ & $+2.0$ \\
yaw-rate tracking & $\exp\!\left(-\lvert\omega_z^c-\omega_z^b\rvert^2/\sigma^2\right),\ \sigma=0.5$ & $+2.0$ \\
feet air time & ${\scriptstyle \mathbb{I}_{\mathrm{move}}\left[\sum_{i\in\mathcal{F}}c_i^{\mathrm{air}}=1\right]\min_{i\in\mathcal{F}}t_i}$ & $+0.5$ \\
feet sliding & $\sum_{i\in\mathcal{F}}c_i\lVert\mathbf{v}^{w}_{i,xy}\rVert_2$ & $-0.4$ \\
base orientation & $\sum_{j\in\{x,y\}}(g_j^b)^2$ & $-3.0$ \\
volume-point penetration & $\sum_j[\delta_j>0](\lVert\dot{\mathbf{p}}_j\rVert_2+10^{-6})\delta_j$ & $-4.0$ \\
joint position limits & ${\scriptstyle \sum_{i\in\mathcal{J}}\left[\operatorname{ReLU}(q_i^{\min}-q_i)+\operatorname{ReLU}(q_i-q_i^{\max})\right]}$ & $-1.0$ \\
joint velocity limits & $\sum_{i\in\mathcal{J}}\operatorname{clip}_{[0,1]}\!\left(\lvert\dot{q}_i\rvert-0.9\dot{q}_i^{\max}\right)$ & $-1.0$ \\
applied torque limits & ${\scriptstyle \sum_{i\in\mathcal{J}}\operatorname{ReLU}\!\left(\lvert\tau_i^a\rvert-0.8\tau_i^{\max}\right)^2}$ & $-0.01$ \\
undesired contacts & $\sum_{i\in\mathcal{J}_{\mathrm{contact}}}[\max_h\lVert\mathbf{F}_{i,h}\rVert_2>1.0]$ & $-1.0$ \\
action rate & $\sum_j(a_{t,j}-a_{t-1,j})^2$ & $-0.01$ \\
\bottomrule
\end{tabularx}
\end{table}
The policy and value networks and the AMP discriminator are optimized with separate objectives. The policy-side update minimizes

\begin{equation}
\begin{aligned}
    \mathcal{L}_{\mathrm{AC}}(\theta,\phi)
    =\; &\mathcal{L}_{\mathrm{ppo}}(\theta)
    + c_1\mathcal{L}_{\mathrm{value}}(\phi)
    + c_2\mathcal{L}_{\mathrm{pred}}(\theta) \\
    &- c_3\mathcal{H}\!\left[\pi_{\theta}(\mathbf{a}_t\mid \mathbf{o}_t,\mathbf{z}_t)\right]
\end{aligned}
\label{eq:actor_critic_loss}
\end{equation}

\begin{table*}[t]
\centering
\caption{\textbf{Performance comparison of different models across various terrains.}}
\label{tab:terrain_performance}
\setlength{\tabcolsep}{2.0pt}
\scriptsize
\resizebox{\textwidth}{!}{
\begin{tabular}{lccccccccccc}
\toprule
Model/Terrain & Boxes & Walk Over Obstacles & Climb Slope & Rough ground & Up Stairs & Climb Down & Down stairs & Climb Up & Gaps Crossing & \textbf{Mean} & \textbf{OOD} \\
\midrule
\textit{(Zhu et al.)}~1-MLP(w/o MoE)~\cite{zhuang2023robot} & \meanstd{57.28}{0.12\%} & \meanstd{66.56}{0.27\%} & \meanstd{81.52}{0.25\%} & \meanstd{94.18}{0.13\%} & \meanstd{4.58}{0.14\%} & \meanstd{94.38}{0.16\%} & \meanstd{0.14}{0.19\%} & \meanstd{21.40}{0.20\%} & \meanstd{0.54}{0.20\%} & \meanstd{46.73}{0.18\%} & \meanstd{21.38}{0.17\%}\\

\textit{(Zhu et al.)}~1-MLP(w/o MoE) (FSM deployment)~\cite{zhu2026hiking} & \meanstd{87.34}{0.23\%} & \meanstd{79.06}{0.31\%} & \meanstd{84.20}{0.15\%} & \meanstd{94.08}{0.24\%} & \meanstd{90.36}{0.12\%} & \meanstd{92.32}{0.21\%} & \meanstd{93.32}{0.13\%} & \meanstd{91.44}{0.29\%} & \meanstd{83.12}{0.29\%} & \meanstd{88.36}{0.22\%} & \meanstd{58.57}{0.16\%}\\

\textit{(Zhu et al.)}~4-MLP(MoE)~\cite{zhu2026hiking} & \meanstd{81.54}{0.14\%} & \meanstd{74.98}{0.29\%} & \meanstd{82.30}{0.28\%} & \meanstd{92.38}{0.20\%} & \meanstd{90.62}{0.26\%} & \meanstd{94.24}{0.27\%} & \meanstd{81.02}{0.26\%} & \meanstd{94.58}{0.30\%} & \meanstd{92.74}{0.12\%} & \meanstd{87.16}{0.24\%} & \meanstd{61.43}{0.19\%}\\
\textit{(Radosavovic et al.)}~Vanilla Transformer~\cite{radosavovic2024humanoid} & \meanstd{89.86}{0.23\%} & \meanstd{84.50}{0.18\%} & \meanstd{85.68}{0.29\%} & \meanstd{94.22}{0.16\%} & \meanstd{89.98}{0.26\%} & \meanstd{94.68}{0.14\%} & \meanstd{90.44}{0.19\%} & \meanstd{94.12}{0.18\%} & \meanstd{90.96}{0.28\%} & \meanstd{90.49}{0.21\%} & \meanstd{84.19}{0.11\%}\\
Ours(\textbf{ParkourFormer}) & \boldmeanstd{91.18}{0.12\%} & \boldmeanstd{86.12}{0.13\%} & \boldmeanstd{95.32}{0.30\%} & \boldmeanstd{96.04}{0.21\%} & \boldmeanstd{94.98}{0.32\%} & \boldmeanstd{95.42}{0.27\%} & \boldmeanstd{95.42}{0.38\%} & \boldmeanstd{94.98}{0.26\%} & \boldmeanstd{96.20}{0.41\%} & \boldmeanstd{93.96}{0.27\%} & \boldmeanstd{93.08}{0.08\%}\\
\bottomrule
\end{tabular}
}
\end{table*}

where $\theta$ and $\phi$ parameterize the actor policy and critic value network, respectively. $\mathcal{L}_{\mathrm{ppo}}(\theta)$ and $\mathcal{L}_{\mathrm{value}}(\phi)$ follow the standard clipped policy and value-function losses in PPO. $\mathcal{L}_{\mathrm{pred}}(\theta)$ is the supervised future-state prediction loss defined in (\ref{eq:pred_loss}), and $\mathcal{H}[\cdot]$ is the entropy regularizer. The coefficients $c_1$, $c_2$, and $c_3$ weight value learning, future-state prediction, and entropy regularization, respectively. The prediction-loss weighting is advantage-dependent: samples with negative advantage receive larger weights, whereas positive-advantage samples retain a weight of 1.0.

The AMP discriminator is optimized separately from the policy-side objective. 
Let $x_\pi$ and $x_E$ denote policy and reference AMP state sequences, respectively. 
We employ a mean-squared objective on the raw scalar discriminator output, with target values $-1$ for policy sequences and $+1$ for reference sequences. 
The discriminator loss, denoted by $\mathcal{L}_{\mathrm{D}}(\psi)$, serves as the AMP loss:
\begin{equation}
\begin{aligned}
    \mathcal{L}_{\mathrm{D}}(\psi)
    ={}&\frac{1}{2}\Big[
    \mathbb{E}_{x_\pi}\!\left(D_\psi(x_\pi)+1\right)^2\\
    &\quad+\mathbb{E}_{x_E}\!\left(D_\psi(x_E)-1\right)^2
    \Big].
\end{aligned}
\label{eq:discriminator_loss}
\end{equation}
where $\psi$ parameterizes the discriminator, and $D_\psi(\cdot)$ denotes the raw discriminator output without a sigmoid layer. 
The discriminator output is used to construct $R_{\mathrm{AMP}}$ and the corresponding PPO returns and advantages, but $\mathcal{L}_{\mathrm{D}}(\psi)$ is not back-propagated through the policy network. 
Thus, the future-prediction head is trained through the policy-side objective, whereas the discriminator follows its separate MSE-based objective.

\Needspace{8\baselineskip}
\section{Experiments}
\label{sec:result}

Following recent perceptive parkour benchmarks, we evaluate ParkourFormer around three central questions:

\textbf{Q1:} Can future-conditioned Transformer policies outperform strong MLP, MoE and Transformer baselines in multi-terrain humanoid parkour?

\textbf{Q2:} Does ParkourFormer generalize across terrains with substantially different geometries, including stairs, gaps, slopes, rough terrain, and obstacles?

\textbf{Q3:} Do the proposed temporal modeling and future prediction mechanisms improve the robustness and stability of unified all-terrain locomotion?

To answer these questions, we evaluate ParkourFormer in both simulation and real-world humanoid experiments, and conduct ablation studies to isolate the contribution of each design component.

\subsection{Environments and Training Details}

All experiments are conducted using the MuJoCo simulation pipeline from Project Instinct~\cite{zhu2026hiking}, with 4,096 parallel environments. Training runs for up to 30,000 iterations on a single NVIDIA RTX 4090D. We repeat each quantitative evaluation with three independent random seeds and report the mean $\pm$ standard deviation.

\subsubsection{Baseline Implementations}
We construct controlled MoE-MLP and Vanilla Transformer baselines under the same training and evaluation protocol as ParkourFormer. The MoE-MLP baseline follows the four-expert architecture of Hiking in the Wild~\cite{zhu2026hiking}, while the Vanilla Transformer baseline follows the causal sequence-modeling formulation of Humanoid Locomotion as Next Token Prediction~\cite{radosavovic2024humanoid}. All baselines use identical observations, rewards, terrain distributions, training budgets, and evaluation environments. These baselines use the same observations, rewards, training budget, and evaluation environments as ParkourFormer. Model size and runtime are summarized in Table~\ref{tab:model_efficiency}.

\begin{table}[tb]
\centering
\caption{\textbf{Model size and runtime.}}
\label{tab:model_efficiency}
\scriptsize
\setlength{\tabcolsep}{2.5pt}
\begin{tabularx}{\linewidth}{@{}>{\raggedright\arraybackslash}X>{\centering\arraybackslash}p{0.14\linewidth}>{\centering\arraybackslash}p{0.14\linewidth}>{\centering\arraybackslash}p{0.14\linewidth}>{\centering\arraybackslash}p{0.14\linewidth}@{}}
\toprule
\textbf{Model} & \textbf{Params. (M)} & \textbf{FLOPs (M)} & \textbf{Train (h)} & \textbf{Infer. (ms)} \\
\midrule
\textit{(Zhuang et al.)}~\cite{zhuang2023robot} \\ \textbf{1-MLP (w/o MoE)} & 1.93 & 2.25 & 22 & 0.16 \\
\textit{(Zhu et al.)}~\cite{zhu2026hiking} \\ \textbf{4-MLP (MoE)} & 3.58 & 3.89 & 50 &  0.28 \\
\textit{(Radosavovic et al.)}~\cite{radosavovic2024humanoid} \\ \textbf{Vanilla Transformer} & 3.63 & 5.68 & 74 & 0.37 \\
(Ours) \\ \textbf{ParkourFormer}  & 3.74 & 7.94 & 89 & 0.53 \\
\bottomrule
\end{tabularx}
\end{table}

\subsubsection{Evaluation Environments}
All simulation and real-world evaluations are conducted on the 29-DoF Unitree G1 humanoid. Simulation evaluation is performed in MuJoCo.
 
\begin{figure*}[t]
    \centering
    \includegraphics[width=\textwidth]{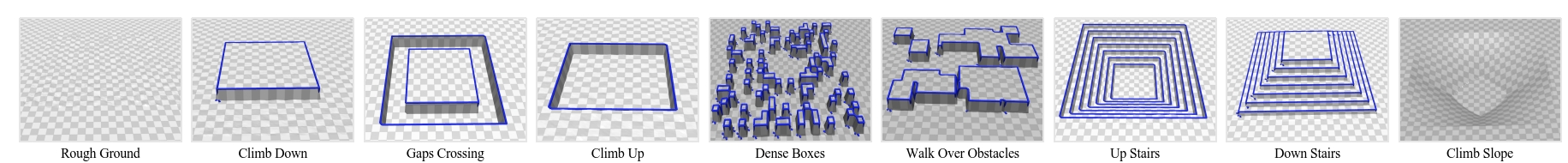}
    \caption{
    \textbf{The training uses a set of nine terrain types.}The Multi-terrain generated based on a unified procedural terrain framework, including scenarios such as Boxes, Walk Over Obstacles, Climb Slope, Rough ground, Up Stairs, Climb Down, Down stairs, Climb Up, and Gaps Crossing. 
    }
    \label{fig:training_terrains}
\end{figure*}

The nine terrain configurations are designed to cover diverse locomotion challenges commonly encountered in real-world traversal, and the training terrains are shown in Fig.~\ref{fig:training_terrains}. Terrain difficulty is varied along three dimensions: geometric variation, such as stair height and slope; obstacle density and spacing; and terrain discontinuity, including gaps and abrupt elevation changes. We evaluate generalization under three regimes: in-distribution (ID) terrains sampled from the training ranges, extrapolated OOD terrains whose geometric parameters exceed the training bounds, and compositional OOD terrains that combine multiple terrain types. The exact settings are listed in Table~\ref{tab:terrain_evaluation}.

\begin{table}[tb]
\centering
\caption{\textbf{Training terrain settings.}}
\label{tab:terrain_training}
\scriptsize
\setlength{\tabcolsep}{3pt}
\renewcommand{\arraystretch}{1.12}
\begin{tabularx}{\linewidth}{@{}>{\raggedright\arraybackslash}p{0.22\linewidth}>{\raggedright\arraybackslash}X>{\raggedright\arraybackslash}p{0.28\linewidth}@{}}
\toprule
\textbf{Terrain category} & \textbf{Attribute} & \textbf{Training range} \\
\midrule
Rough terrain & Perlin-noise amplitude & $0.00$--$0.10$ \\

Gap crossing & Gap distance (m) & $0.10$--$0.60$ \\
 & Gap depth (m) & $0.40$--$0.60$ \\

Stairs & Single-step height (m) & $0.05$--$0.40$ \\

Box obstacles & Obstacle height (m) & $0.05$--$0.45$ \\
 & Obstacle width/length (m) & $0.80$--$1.50$ \\

Randomized box & Mean box height (m) & $0.10$--$0.40$ \\
 & Perturbed box height (m) & $0.05$--$0.45$ \\

Slope & Slope parameter (rad) & $0.00$--$0.60$ \\
\bottomrule
\end{tabularx}
\end{table}


\begin{figure*}[htbp]
    \centering
    \includegraphics[width=\textwidth]{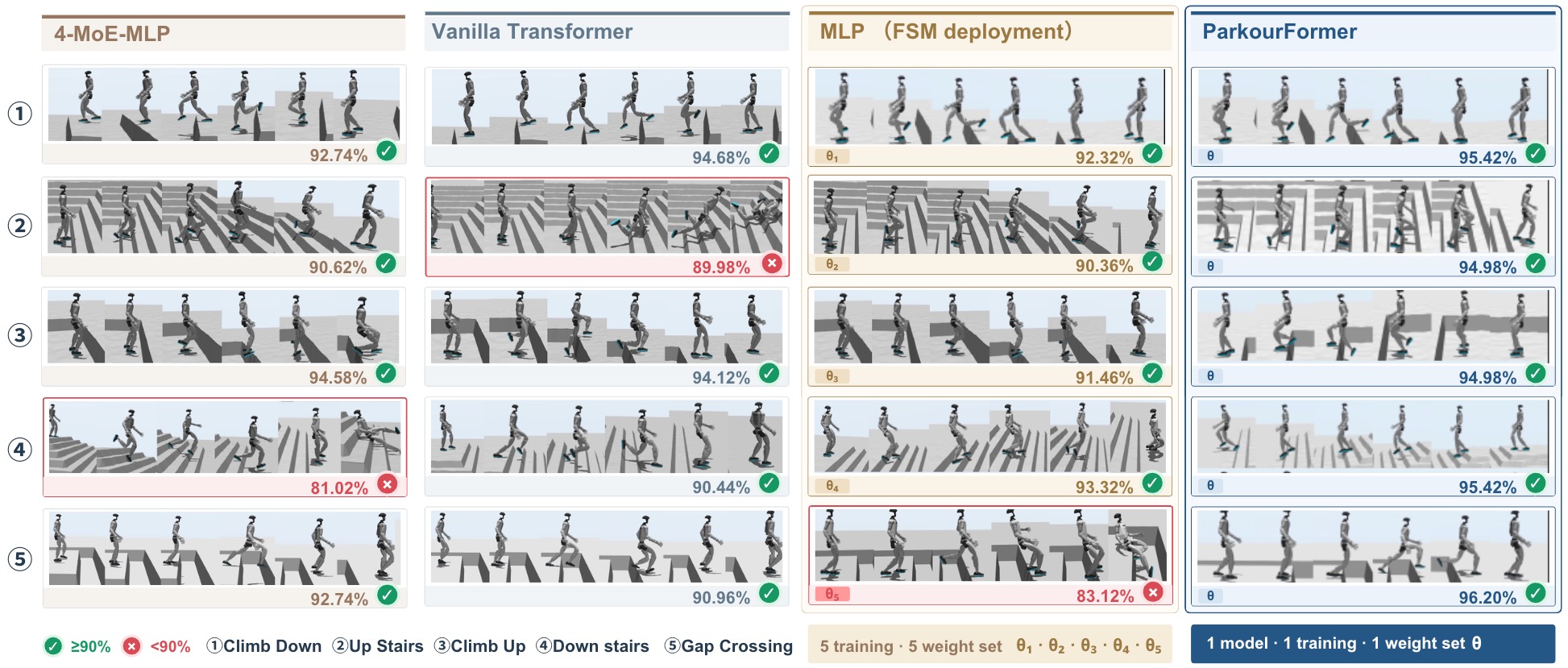}
    \caption{
    \textbf{Comparison of multi-terrain locomotion strategies.}
    ParkourFormer jointly learns all five representative terrains using one model, one training run, and one shared parameter set $\theta$.
    In contrast, terrain-specific MLP/MoE baselines require separate training runs and parameter sets $\{\theta_1,\ldots,\theta_5\}$, which are selected during deployment by an FSM (finite-state machine).
    Although FSM-based switching can improve terrain-wise specialization, it adds deployment complexity and may introduce unstable transitions in real-world parkour, increasing the risk of failure or falls.
    Check/cross symbols indicate success rates above/below 90\%.
    }
    \label{fig:multi-terrain}
\end{figure*}

\begin{table}[h]
\centering
\caption{\textbf{Evaluation terrain settings across three generalization regimes.}
The gap depth is fixed at $0.5$~m for mixed terrains.}
\label{tab:terrain_evaluation}
\scriptsize
\setlength{\tabcolsep}{3pt}
\renewcommand{\arraystretch}{1.12}
\begin{tabularx}{\linewidth}{@{}>{\raggedright\arraybackslash}p{0.19\linewidth}>{\raggedright\arraybackslash}X>{\raggedright\arraybackslash}p{0.30\linewidth}@{}}
\toprule
\textbf{Terrain} & \textbf{Attribute} & \textbf{Evaluation value} \\
\midrule
\multicolumn{3}{@{}l}{\textbf{(a) In-distribution (ID):} Sampled within the Table~\ref{tab:terrain_training}.}\\
\cmidrule{1-3}
All nine terrains &  &  Ranges in Table~\ref{tab:terrain_training}\\
\midrule
\multicolumn{3}{@{}l}{\textbf{(b) Simulation Extrapolated OOD:} $1.1\times$ the training upper bounds.} \\
\cmidrule{1-3}
Rough terrain & Perlin-noise amplitude & $0.11$ \\
Gap crossing & Gap distance / depth (m) & $0.66$ \\
Stairs & Single-step height (m) & $0.44$ \\
Box obstacles & Obstacle height (m) & $0.50$ \\
 & Obstacle width/length (m) & $1.65$ \\
Random boxes & Mean / perturbed height (m) & $0.44$ / $0.50$ \\
Slope & Slope parameter (rad) & $0.66$ \\
\midrule
\multicolumn{3}{@{}l}{\textbf{(c) Compositional OOD(Real Deployment Only):} Mixed terrains.} \\
\cmidrule{1-3}
Mix-L1 & Stair height (m) / slope (rad) / gap (m) & $0.10$ / $0.12$ / $0.20$ \\
Mix-L2 & Stair height (m) / slope (rad) / gap (m) & $0.24$ / $0.30$ / $0.36$ \\
Mix-L3 & Stair height (m) / slope (rad) / gap (m) & $0.38$ / $0.48$ / $0.52$ \\
\bottomrule
\end{tabularx}
\end{table}

\subsection{Multi-terrain Performance}

As shown in Table~\ref{tab:terrain_performance} and Fig.~\ref{fig:multi-terrain}, ParkourFormer achieves the best overall performance, with a mean success rate of 93.96\%, compared with 90.49\% for the Vanilla Transformer and 87.16\% for the MoE-MLP baseline. All controlled baselines are jointly trained over the same terrain distribution with a single set of policy weights. Under this setting, the 1-MLP shows strong interference across heterogeneous motion patterns, especially on dynamic behaviors such as stair descent. Training a separate 1-MLP policy for each terrain and using an FSM to switch among the corresponding weight sets improves the mean success rate from 46.73\% to 88.36\%, but requires multiple training runs and achieves only 58.57\% OOD success. We further observe that policy switching can introduce motion discontinuities and falls near terrain transitions. In contrast, ParkourFormer learns all terrains jointly in one training process and achieves 93.96\% ID and 93.08\% OOD success with one shared parameter set. The Vanilla Transformer comparison further indicates that the gain is not solely due to the Transformer backbone.

\begin{figure}[h]
    \centering
    \includegraphics[width=\linewidth]{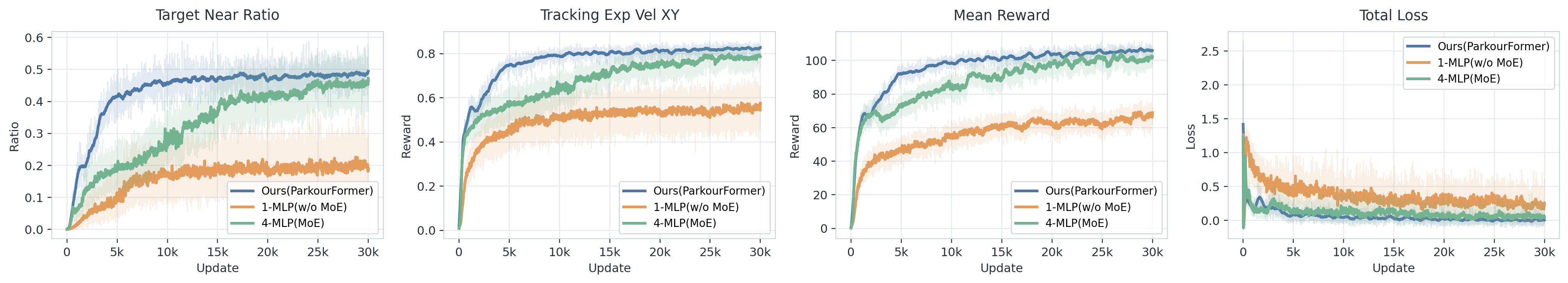}
    \caption{
    \textbf{Training dynamics comparison across different policy architectures.}
    ParkourFormer reaches higher task performance with less iterations in training than the compared baselines, while exhibiting smoother learning dynamics.
    }
    \label{fig:training_dynamics}
\end{figure}

The comparison with the Vanilla Transformer further indicates that the improvement cannot be explained solely by adopting a Transformer backbone. ParkourFormer also achieves the highest Target Near Ratio and Tracking Velocity scores in Table~\ref{tab:quantitative_comparison}, indicating improved target reaching and velocity-command tracking. Fig.~\ref{fig:training_dynamics} further shows that ParkourFormer reaches higher task performance earlier during training, suggesting that predictive supervision provides an additional learning signal for unified multi-terrain optimization. The contributions of future-state prediction, predictive supervision, and downstream future conditioning are further examined through controlled ablations in Sec.~B.

\begin{table}[h]
\centering
\caption{\textbf{Quantitative comparison results.}}
\label{tab:quantitative_comparison}
\small
\resizebox{\linewidth}{!}{%
\begin{tabular}{lcc}
\toprule
 & \multicolumn{1}{c}{\textbf{Target Near Ratio} } & \multicolumn{1}{c}{\textbf{Tracking Vel} } \\
\cmidrule(lr){2-2}\cmidrule(lr){3-3}
\textbf{Model} & \textbf{Score} & \textbf{Score} \\
\midrule
\textit{(Zhuang et al.)}~1-MLP(w/o MoE)~\cite{zhuang2023robot} & 0.199 & 0.554 \\
\textit{(Zhu et al.)}~4-MLP(MoE)~\cite{zhu2026hiking} & 0.437 & 0.793 \\
\textit{(Radosavovic et al.)}~Vanilla Transformer~\cite{radosavovic2024humanoid} & 0.462 & 0.812 \\
Ours(ParkourFormer) & \textbf{0.489} & \textbf{0.837} \\
\bottomrule
\end{tabular}}
\end{table}





\begin{table*}[t]
\centering
\caption{\textbf{Ablation results across terrain tasks.}}
\label{tab:ablation_results}
\setlength{\tabcolsep}{3.0pt}
\scriptsize
\resizebox{\textwidth}{!}{
\begin{tabular}{lcccccccccc}
\toprule
Model/Terrain & Boxes & Walk Over Obstacles & Climb Slope & Rough ground & Up Stairs & Climb Down & Down stairs & Climb Up & Gaps Crossing & \textbf{Mean}\\
\midrule
Ours(\textbf{ParkourFormer}) & \boldmeanstd{91.18}{0.12\%} & \boldmeanstd{86.12}{0.13\%} & \boldmeanstd{95.32}{0.30\%} & \boldmeanstd{96.04}{0.21\%} & \boldmeanstd{94.98}{0.32\%} & \boldmeanstd{95.42}{0.27\%} & \boldmeanstd{95.42}{0.38\%} & \boldmeanstd{94.98}{0.26\%} & \boldmeanstd{96.20}{0.41\%} & \boldmeanstd{93.96}{0.27\%}\\
w/o\ supervise signal(MSE) & \meanstd{85.48}{0.22\%} & \meanstd{83.42}{0.25\%} & \meanstd{95.10}{0.19\%} & \meanstd{93.06}{0.14\%} & \meanstd{93.78}{0.27\%} & \meanstd{95.24}{0.38\%} & \meanstd{89.96}{0.17\%} & \boldmeanstd{94.98}{0.20\%} & \meanstd{95.42}{0.42\%} & \meanstd{91.83}{0.21\%}\\
w/o\ RGB-D Query & \meanstd{86.94}{0.19\%} & \meanstd{79.74}{0.15\%} & \meanstd{85.56}{0.20\%} & \meanstd{90.74}{0.19\%} & \meanstd{92.60}{0.42\%} & \meanstd{94.40}{0.33\%} & \meanstd{90.68}{0.25\%} & \meanstd{75.84}{0.36\%} & \meanstd{24.24}{0.22\%} & \meanstd{80.08}{0.26\%}\\
w/o\ future prediction & \meanstd{89.64}{0.28\%} & \meanstd{85.98}{0.19\%} & \meanstd{92.70}{0.22\%} & \meanstd{94.02}{0.21\%} & \meanstd{94.18}{0.39\%} & \meanstd{95.24}{0.39\%} & \meanstd{93.52}{0.21\%} & \meanstd{94.82}{0.38\%} & \meanstd{95.04}{0.43\%} & \meanstd{92.79}{0.30\%}\\
Action Without Prediction & \meanstd{89.40}{0.27\%} & \meanstd{83.20}{0.26\%} & \meanstd{89.80}{0.14\%} & \meanstd{94.60}{0.28\%} & \meanstd{92.80}{0.32\%} & \meanstd{92.40}{0.31\%} & \meanstd{91.80}{0.34\%} & \meanstd{93.20}{0.36\%} & \meanstd{93.20}{0.29\%} & \meanstd{91.16}{0.28\%}\\
1-Frame Prediction & \meanstd{90.80}{0.11\%} & \meanstd{81.60}{0.15\%} & \meanstd{88.40}{0.16\%} & \meanstd{93.60}{0.23\%} & \meanstd{93.00}{0.42\%} & \meanstd{94.60}{0.25\%} & \meanstd{92.60}{0.25\%} & \meanstd{94.60}{0.42\%} & \meanstd{92.60}{0.33\%} & \meanstd{91.31}{0.26\%}\\
4-Frame Prediction & \meanstd{90.60}{0.28\%} & \meanstd{77.00}{0.23\%} & \meanstd{70.60}{0.15\%} & \meanstd{95.20}{0.20\%} & \meanstd{90.60}{0.27\%} & \meanstd{95.20}{0.44\%} & \meanstd{90.60}{0.35\%} & \meanstd{93.80}{0.38\%} & \meanstd{93.00}{0.26\%} & \meanstd{88.51}{0.28\%}\\
8-Frame Prediction & \meanstd{77.40}{0.14\%} & \meanstd{72.60}{0.13\%} & \meanstd{52.40}{0.10\%} & \meanstd{95.20}{0.30\%} & \meanstd{92.20}{0.24\%} & \meanstd{95.00}{0.41\%} & \meanstd{0.20}{0.12\%} & \meanstd{94.00}{0.37\%} & \meanstd{93.80}{0.28\%} & \meanstd{74.76}{0.23\%}\\
AMP Without Prediction & \meanstd{88.80}{0.27\%} & \meanstd{75.40}{0.10\%} & \meanstd{71.80}{0.18\%} & \meanstd{95.20}{0.15\%} & \meanstd{92.80}{0.44\%} & \meanstd{92.08}{0.33\%} & \meanstd{88.20}{0.22\%} & \meanstd{93.80}{0.31\%} & \meanstd{86.60}{0.25\%} & \meanstd{87.19}{0.25\%}\\
\bottomrule
\end{tabular}
}
\end{table*}

\subsection{Ablation Experiments}

Table~\ref{tab:ablation_results} examines the role of predictive supervision, downstream future-state conditioning, and prediction horizon. Removing the future-prediction pathway reduces the mean success rate from 93.96\% to 92.79\%, whereas retaining the prediction pathway without supervision causes a much larger drop to 91.83\%. Without the prediction loss, the predicted states are no longer constrained to represent meaningful future proprioceptive targets while still influencing downstream modules, leading to degraded control. To distinguish this effect from generic auxiliary regularization, we retain supervised prediction but remove its output from the action head; performance drops to 91.16\%, showing that the predicted states provide useful information beyond supervising the shared backbone. Removing them from the AMP input further reduces performance to 87.19\%, indicating an additional benefit for temporal motion regularization. Finally, two-frame prediction performs best: one frame provides less anticipatory context, while four- and eight-frame prediction reduce mean success to 88.51\% and 74.76\%, respectively, reflecting the increasing difficulty of prediction.

\subsection{Real-world Experiments}
\label{sec:real_world}

\subsubsection{Hardware, Perception, and Control}
We deploy ParkourFormer on a Unitree G1 humanoid with 29 degrees of freedom. An Intel RealSense D435i supplies the exteroceptive depth observations used by the terrain encoder. The policy executes at $50$~Hz, matching the control frequency in simulation. At every control step, the policy outputs a 29-dimensional action delta around the nominal pose; this output is converted into joint-position targets and tracked by the robot's low-level PD controller.

\subsubsection{Real-world Success Rate and Generalization}

The mixed-terrain assemblies employed in the real-world evaluation are not encountered during training and are therefore evaluated as out-of-distribution test conditions. Their configurations are listed in Table~\ref{tab:terrain_evaluation}. Figure~\ref{fig:real_world_deployment} provides representative deployment snapshots: panels (a) and (b) correspond to the Mix-L2 assembly, whereas panels (c) and (d) show deployment on outdoor step terrains. We additionally include four natural outdoor OOD terrain categories in Table~\ref{tab:real_world_mixed_terrain}; their success-rate entries are left blank until the corresponding trials are recorded. These images document the evaluation conditions qualitatively; the corresponding task success rates are reported in Table~\ref{tab:real_world_mixed_terrain}. The experimental results show that ParkourFormer maintains stable locomotion on these unseen terrains, indicating a certain degree of OOD generalization capability.

\subsubsection{Sim-to-Real Gap and Mitigation}
At deployment, we used the same depth representation and observation normalization as during training. Additionally, we stochastically perturbed the contact attributes, actuator responses, observation noise, depth corruption, and camera extrinsics throughout the training phase. 

\begin{center}
\includegraphics[width=\linewidth]{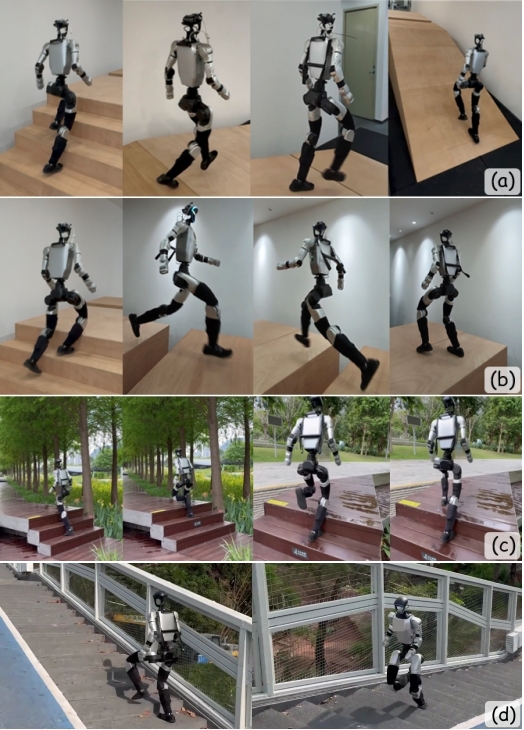}
\captionof{figure}{\textbf{Representative real-world deployments.} Panels (a) and (b) show the Unitree G1 traversing the Mix-L2 mixed-terrain assembly described in Table~\ref{tab:terrain_evaluation}; panels (c) and (d) show deployment on outdoor step terrains.}
\label{fig:real_world_deployment}
\end{center}

\begin{center}
\captionof{table}{\textbf{Real-world task success rates.}}
\label{tab:real_world_mixed_terrain}
\footnotesize
\setlength{\tabcolsep}{4pt}
\renewcommand{\arraystretch}{1.10}
\begin{tabularx}{\linewidth}{@{}>{\raggedright\arraybackslash}p{0.25\linewidth}>{\centering\arraybackslash}X>{\centering\arraybackslash}p{0.26\linewidth}@{}}
\toprule
\textbf{Terrain} & \textbf{Settings} & \textbf{Success Rate} \\
\midrule
\multicolumn{3}{c}{\textit{Mixed-Terrain Settings}} \\
\midrule
Mix-L1 & Stairs--Gap--Slope & $100\%\,{\scriptstyle(20/20)}$ \\
Mix-L2 & Stairs--Gap--Slope & $95\%\,{\scriptstyle(19/20)}$ \\
Mix-L3 & Stairs--Gap--Slope & $80\%\,{\scriptstyle(16/20)}$ \\
\midrule
\multicolumn{3}{c}{\textit{Single-Terrain Settings}} \\
\midrule
Up Stairs & $0.32$~m & $95\%\,{\scriptstyle(19/20)}$ \\
Down stairs & $0.32$~m & $95\%\,{\scriptstyle(19/20)}$ \\
Gaps Crossing & $0.50$~m & $90\%\,{\scriptstyle(18/20)}$ \\
\midrule
\multicolumn{3}{c}{\textit{OOD Terrain Settings}} \\
\midrule
Outdoor Stairs & Natural outdoor staircase & $95\%\,{\scriptstyle(19/20)}$ \\
Long Stairs & Extended staircase & $80\%\,{\scriptstyle(16/20)}$ \\
Wild Grassland & Unstructured grassy terrain & $100\%\,{\scriptstyle(20/20)}$ \\
\bottomrule
\end{tabularx}
\end{center}

\section{Conclusion}
\label{sec:conclusion}

We presented ParkourFormer, a Transformer-based framework that formulates humanoid parkour as a future-conditioned sequence modeling problem. By querying historical sensorimotor trajectories, ParkourFormer predicts short-horizon future proprioceptive states under explicit supervision and uses them to condition both action generation and motion regularization. This design enables heterogeneous locomotion behaviors to be jointly learned in a single training process and deployed with one shared set of policy weights. Experiments on the Unitree G1 show consistent improvements over controlled MLP, MoE-MLP, and Vanilla Transformer baselines across diverse parkour terrains. Ablations further demonstrate that the gains arise from the combination of temporal sequence modeling, predictive supervision, future-conditioned control, and future-aware AMP regularization, rather than increased model capacity alone. Overall, ParkourFormer shows that explicitly structuring temporal representations around short-horizon state prediction provides an effective approach to unified multi-terrain humanoid locomotion. Future work will focus on improving sim-to-real robustness and extending the framework to more dynamic whole-body parkour behaviors.



\bibliographystyle{bib/IEEEtran}
\bibliography{bib/parkourformer}

\end{document}